\documentclass[pmlr]{jmlr}

\RequirePackage{graphicx}
 \usepackage{booktabs}
\usepackage{longtable}
\makeatletter
\def\set@curr@file#1{\def\@curr@file{#1}} 
\makeatother
\usepackage[load-configurations=version-1]{siunitx} 

\theorembodyfont{\upshape}
\theoremheaderfont{\scshape}
\theorempostheader{:}
\theoremsep{\newline}

\jmlrvolume{}
\jmlryear{}
\jmlrworkshop{}

\title[Localising the SOZ from SPES Responses with a Transformer]{Localising the Seizure Onset Zone from Single-Pulse Electrical Stimulation Responses with a CNN Transformer}

\author{
\Name{Jamie Norris}\textsuperscript{1,2},
\Name{Aswin Chari}\textsuperscript{3,4},
\Name{Dorien van Blooijs}\textsuperscript{5,6},
\Name{Gerald Cooray}\textsuperscript{7},
\Name{Karl Friston}\textsuperscript{2},
\Name{Martin Tisdall}\textsuperscript{3,4},
\Name{Richard Rosch}\textsuperscript{2,8,9} \\
\addr 
\textsuperscript{1}UCL Institute of Health Informatics, London, UK \\
\textsuperscript{2}Wellcome Centre for Human Neuroimaging, UCL Queen Square Institute of Neurology, London, UK \\
\textsuperscript{3}Department of Neurosurgery, Great Ormond Street Hospital, London, UK \\
\textsuperscript{4}Developmental Neuroscience, UCL Institute of Child Health, London, UK \\
\textsuperscript{5}UMC Utrecht Brain Center, Department of Neurology and Neurosurgery, University Medical Center Utrecht, Utrecht, The Netherlands \\
\textsuperscript{6}Stichting Epilepsie Instellingen Nederland (SEIN), Hoofddorp, The Netherlands \\
\textsuperscript{7}Department of Neurophysiology, Great Ormond Street Hospital, London, UK \\
\textsuperscript{8}Departments of Neurology and Pediatrics, Columbia University, New York, NY, USA \\
\textsuperscript{9}Department of Neurophysiology, King's College Hospital NHS Foundation Trust, London, UK
}

\begin{document}

\maketitle

\begin{abstract}

Epilepsy is one of the most common neurological disorders, often requiring surgical intervention when medication fails to control seizures. For effective surgical outcomes, precise localisation of the epileptogenic focus -- often approximated through the Seizure Onset Zone (SOZ) -- is critical yet remains a challenge. Active probing through electrical stimulation is already standard clinical practice for identifying epileptogenic areas. Our study advances the application of deep learning for SOZ localisation using Single-Pulse Electrical Stimulation (SPES) responses, with two key contributions. Firstly, we implement an existing deep learning model to compare two SPES analysis paradigms: \textit{divergent} and \textit{convergent}. These paradigms evaluate outward and inward effective connections, respectively. We assess the generalisability of these models to unseen patients and electrode placements using held-out test sets. Our findings reveal a notable improvement in moving from a divergent (AUROC: 0.574) to a convergent approach (AUROC: 0.666), marking the first application of the latter in this context. Secondly, we demonstrate the efficacy of CNN Transformers with cross-channel attention in handling heterogeneous electrode placements, increasing the AUROC to 0.730\footnote{\url{https://github.com/norrisjamie23/Localising_SOZ_from_SPES}}. These findings represent a significant step in modelling patient-specific intracranial EEG electrode placements in SPES. Future work will explore integrating these models into clinical decision-making processes to bridge the gap between deep learning research and practical healthcare applications.
\end{abstract}

\section{Introduction}
\label{sec:intro}

Epileptic seizures originate from diverse brain networks, exhibiting highly individualised, time-varying interictal and ictal patterns that often require uniquely personalised treatments to achieve complete seizure control. Around 1 in 100 people will be diagnosed with epilepsy \citep{vaughan2018estimation}, and approximately one-third of cases are resistant to medication \citep{kalilani2018epidemiology}. Many of these patients are candidates for surgery, which is a safe and effective treatment option to reduce or altogether eliminate seizures \citep{wiebe2001randomized}. This approach entails identifying and removing focal brain regions implicated in seizure generation, known as the epileptogenic zone. The seizure onset zone (SOZ) is the primary clinically identifiable proxy for these critical areas \citep{jehi2018epileptogenic}. However, identifying the SOZ requires intracranial recordings of spontaneously occurring seizures. Seizures are comparatively rare events and often require extended hospital stays to capture, incurring high costs and stress for patients. Even with the current gold standard evaluation, outcomes remain comparatively poor in cases requiring invasive intracranial evaluation. Moreover, a putative SOZ is not identified at all in approximately 15\% of cases, preventing surgical intervention \citep{uk2021uk}.

Rather than solely depending on the passive recording of spontaneous seizures, actively probing the epileptic brain's circuitry is also part of current clinical standard practice. One such method is single-pulse electrical stimulation (SPES), in which an isolated, brief electrical stimulus is applied to individual intracranial electrode contacts \citep{matsumoto2017single}, and evoked distributed responses are measured across the remaining contacts. SPES serves as a proactive tool in the presurgical evaluation of epilepsy, enabling the testing of specific hypotheses about the excitability of targeted regions, with some correlation with their epileptogenicity.

SPES elicits two primary types of responses: early and delayed. Early responses, emerging within 100ms post-stimulation and time-locked to the stimulus, are termed cortico-cortical evoked potentials (CCEPs) (Figure \ref{fig:paradigms}D). These CCEPs signify directed or effective connectivity; for instance, if site A is stimulated, a CCEP observed at site B indicates directed neural propagation from A to B. While CCEPs occur in epileptogenic and non-epileptogenic tissue, their characteristics reflect the proximity of the stimulation and recording sites to the SOZ \citep{hays2023cortico}. Given their time-locked nature, CCEPs are typically averaged across 10-50 trials to enhance their signal-to-noise ratio.


In contrast, delayed responses, which occur between 100ms and 1s post-stimulation and resemble epileptiform discharges, reflect focal cortical irritability or even epileptogenicity (Figure \ref{fig:paradigms}B). These responses have been reported in patients with temporal and frontal lobe epilepsies, and the removal of sites exhibiting these is associated with overall favourable outcomes \citep{valentin2005single}. Unlike the consistent and time-locked nature of CCEPs, delayed responses are neither consistent across trials nor time-locked to the stimulus, meaning they are less apparent when averaging across trials. Furthermore, other forms of inter-trial variability may also reflect epileptogenicity \citep{cornblath2023quantifying}.

\begin{figure}[!h]
\floatconts
  {fig:paradigms}
  {\caption{\textbf{SPES analysis paradigms.} \textbf{A:} Example stimulation and recording sites under a \textit{divergent} paradigm. Example responses are shown in \textbf{B}. The same is shown for a \textit{convergent} paradigm in \textbf{C} and \textbf{D}. A delayed response is shown with an arrow in \textbf{B}, and an early response (ER) is shown in \textbf{D}. For \textbf{B} and \textbf{D}, $A \rightarrow B$ corresponds to the response at electrode B upon stimulation at A.}}
  {\includegraphics[width=0.75\linewidth]{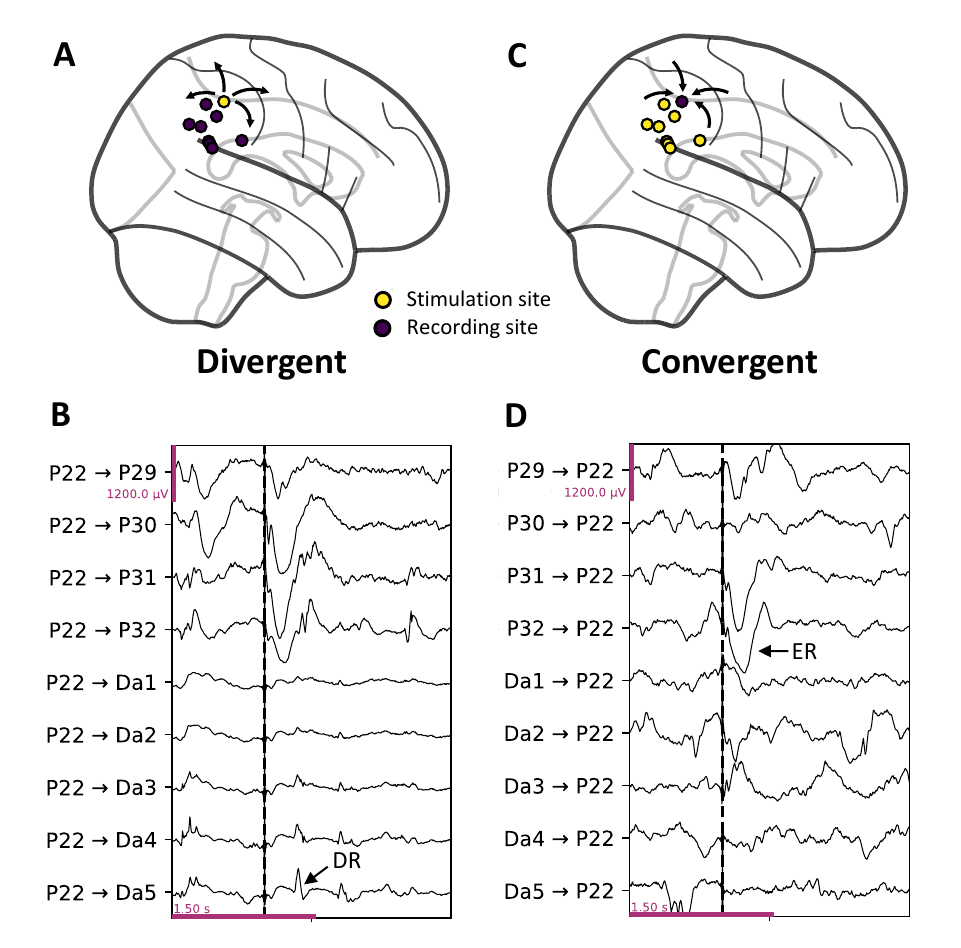}}
\end{figure}

Recent advances in machine learning, particularly in deep learning, are well-positioned to model complex multidimensional time-series data, such as those generated by SPES. This capability makes deep learning methods particularly promising for integrating multi-channel SPES responses to localise the SOZ. However, determining the most effective approach to leverage these advanced techniques remains a significant challenge due to particular data features, such as patient-specific electrode placements and variability in the background EEG activity. 

In this work, we first address the question of the appropriate analysis paradigm for multi-channel SPES data. An individual node $S$ can be classified based on its outward connections, i.e., how other sites respond to stimulation at $S$. This \textit{divergent} paradigm \citep{miller2021basis} is commonplace and has previously been employed for SOZ localisation \citep{johnson2022localizing}. Alternatively, individual nodes can be classified based on their inward connections, i.e., how they respond to stimulation at other sites. This \textit{convergent} paradigm is particularly relevant for identifying delayed responses since these are associated with epileptogenic response sites rather than epileptogenic stimulation sites. This approach aligns with findings from \cite{hays2023cortico}, where responses at SOZ sites upon stimulation at non-SOZ sites were more significant than vice-versa, especially at higher currents, suggesting that the \textit{convergent} paradigm could provide a more effective means for SOZ localisation.

Secondly, effectively modelling patient-specific and heterogeneous multi-channel data presents a significant challenge. Most machine learning applications to time-series data assume a fixed number of channels with consistent ordering. However, intracranial EEG data varies in channel numbers and placements across patients, impeding model generalisability. Therefore, developing methods that can accommodate the unique and complex nature of patient-specific intracranial EEG datasets is an essential issue that needs to be addressed.

This paper introduces a CNN Transformer with cross-channel attention aimed at enhancing generalisability to unseen patients and managing the complexities of intracranial EEG data. Our study focuses on:

\begin{itemize}
\item Adapting an existing deep learning model to compare \textit{divergent} and \textit{convergent} paradigms for SOZ localisation, revealing that switching to a convergent approach improved the AUROC from 0.574 to 0.666.
\item Implementing and evaluating CNN Transformer models, outperforming existing methods with an AUROC of 0.730. We investigate the relationship between model performance and surgical outcome.
\end{itemize}

\subsection*{Generalisable Insights about Machine Learning in the Context of Healthcare}

There is an increasing emphasis on identifying interictal biomarkers of the epileptogenic zone. One such approach is through stimulation techniques used to probe the epileptic brain, and deep learning is well suited to model the resulting multidimensional time-series data. In this work, we consider Single-Pulse Electrical Stimulation, focusing on two aspects: how these multi-channel responses to SPES can be represented and the best way to model heterogeneous intracranial electrode numbers and placements.

Our findings underscore the potential of deep learning to improve diagnostic approaches and treatment strategies in epilepsy. Specifically, we demonstrate the model's capability to generalise to unseen patients within the same centre, setting a foundation for future research to extend this capability across diverse clinical environments and electrode configurations. This work holds broader implications for neurological research, particularly in contexts where precise localisation of brain activity is essential.


\section{Related Work}
\label{sec:related_work}

\subsection{Localising the SOZ from SPES responses}
\label{sec:soz_localisation}
\cite{malone2022machine} evaluated various machine learning algorithms for localising the SOZ from SPES responses, classifying single-channel responses based on whether the recording electrode was in the SOZ. The most effective methods included a 1D convolutional neural network, a gradient-boosted decision tree, and an ensemble approach. In a different study, \cite{yang2024localizing} focused on computing various CCEP features to classify both stimulated and recording electrodes as SOZ/non-SOZ. These methods, however, consider each channel in isolation, leading to multiple predictions per channel and thereby not integrating the multidimensional data features into the classification. In contrast, \cite{johnson2022localizing} adopted a multi-channel approach using a divergent paradigm with a ResNet convolutional neural network. This method classifies the stimulated stereo-EEG (SEEG) channel based on responses from other channels. A limitation of this approach is that the network architecture chosen for this analysis requires a consistent number of channels and associates parameters with specific channel positions. To mitigate this, the authors randomly selected a subset of 40 channels for each pass, sorting these channels based on their proximity to the stimulated electrode. This method's generalisability was validated using leave-one-out cross-validation. While this approach is suited to extracting temporal features, its effectiveness could benefit from adopting a more sophisticated method to focus on important response channels. Rather than having parameters associated with each of a fixed number of channels, capturing a context-aware representation across all channels may deal better with the heterogeneity of patient-specific seizure networks. Furthermore, this fixed channel count necessitates balancing the model's prediction accuracy and generalisability.

\subsection{Transformers for Intracranial EEG}
\label{sec:transformers}
Transformers are a class of deep learning architecture built upon the self-attention mechanism \citep{vaswani2017attention, bahdanau2014neural}. Initially proposed in the context of natural language processing (NLP), Transformers have since been applied in a range of domains, including computer vision \citep{dosovitskiy2020image}, reinforcement learning \citep{chen2021decision}, and time series \citep{lim2021temporal}. In NLP, attention mechanisms efficiently encode sequential information across word tokens. This concept is extendable to time series analysis, where attention can instead be applied across temporal sequences. In the context of epilepsy, Transformers have predominantly been applied to electrophysiological data for seizure detection and prediction \citep{hussein2022multi}.

Beyond modelling temporal dependencies, Transformers can compute attention between channels, treating each EEG electrode as a token to capture a global context. Encoder-only models, such as BERT \citep{devlin2018bert} and the Vision Transformer \citep{dosovitskiy2020image}, offer flexibility in handling a variable number of tokens due to the utilisation of a class token that represents aggregated information across all tokens. These characteristics may be advantageous in extracting features across channels in intracranial EEG, where the number and placement of electrodes are highly heterogeneous across patients. 

An illustrative example is provided by Brant, a foundational model explicitly developed for the analysis of intracranial EEG \citep{zhang2024brant}. This model integrates spatial (cross-channel) and temporal attention mechanisms, enabling it to generalise effectively across various downstream tasks, including neural signal forecasting and seizure detection. However, it remains an open question whether such approaches can accurately model evoked responses, such as those generated by SPES.

\section{Methods}
\label{sec:methods}

\subsection{Notation}
To describe each network, we use the following notation. The output for a given stimulation trial is denoted as $x_d \in \mathbb{R}^{N \times C_{d,i} \times T}$ for the divergent paradigm, and $x_c \in \mathbb{R}^{N \times C_{c,i} \times T}$ for the convergent paradigm. Here, $N$ represents the number of trials, $C_{d,i}$ (or $C_{c,i}$ for a convergent paradigm) indicates the number of channels for patient $i$, and $T$ is the number of time points. Averaging these responses across trials results in $\bar{x} \in \mathbb{R}^{C_{i} \times T}$, which enhances the signal-to-noise ratio and facilitates a single prediction per electrode.

\subsection{Models}
We compare the following model architectures:
\begin{enumerate}
    \item \textbf{CNN\textsubscript{divergent}} \\
    A multi-scale 1D variant \citep{wang2018csi} of a ResNet convolutional neural network \citep{he2016deep}\footnote{\url{https://github.com/geekfeiw/Multi-Scale-1D-ResNet}}.
    \item \textbf{CNN\textsubscript{convergent}} \\
    The same architecture as CNN\textsubscript{divergent}, but using a \textit{convergent} paradigm instead of a \textit{divergent} paradigm.
    \item \textbf{CNN Transformer} \\
    An encoder-only Transformer \citep{vaswani2017attention}. A CNN is applied to each channel, and cross-channel attention extracts features across channels. A \textit{convergent} paradigm was used.
\end{enumerate}


 \begin{figure}[!h]
\floatconts
  {fig:models}
  {\caption{\textbf{Network architectures.} \textbf{Left:} CNN. A ResNet creates a cross-channel embedding, and a multi-layer perceptron (MLP) classifies this as being in or out of the Seizure Onset Zone (SOZ). \textbf{Right:} CNN Transformer. First, a CNN is independently applied to each channel to extract per-channel features. These embeddings are then subjected to cross-channel attention before an MLP is used to classify the electrode as SOZ/non-SOZ.}}
  {\includegraphics[width=0.75\linewidth]{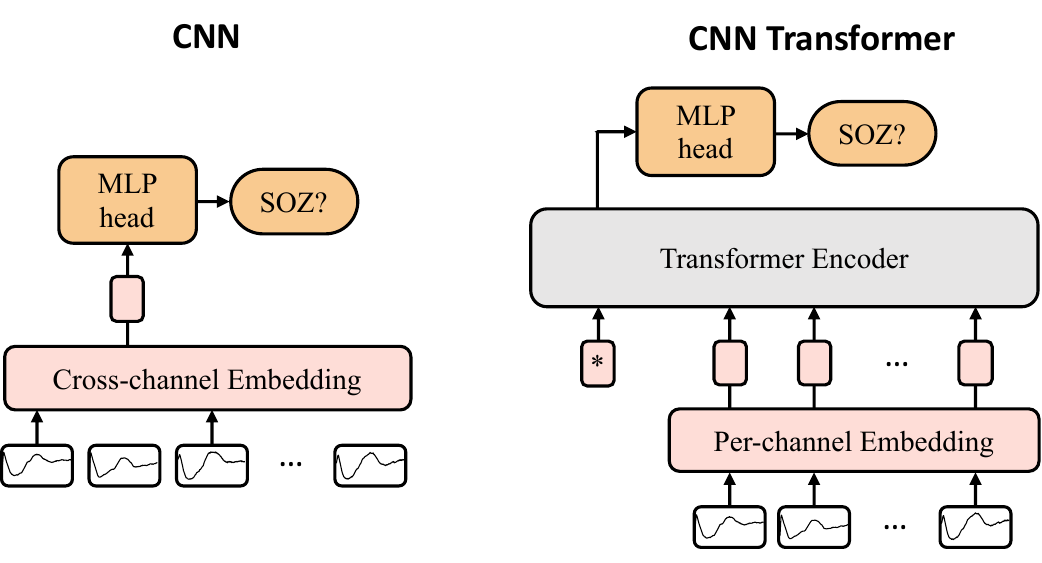}}
\end{figure}

\subsubsection[CNN (divergent)]{CNN\textsubscript{divergent}}
To establish a baseline in our study, we first implemented the method previously applied by \cite{johnson2022localizing} for the same task. Specifically, the authors used a multi-scale 1D ResNet convolutional neural network. ResNets help mitigate overfitting through residual connections, and the multi-scale architecture allows the model to learn features at various temporal scales.

In their study, \cite{johnson2022localizing} adopted a divergent paradigm, where the objective is to classify a stimulating electrode based on the evoked responses, $x_d$, at other electrodes. As described in section \ref{sec:soz_localisation}, a random subset of 40 channels were selected during each training and inference pass and arranged in ascending order of distance from the stimulating electrode.

Two modifications were implemented in this approach. Firstly, the number of input channels was treated as a hyperparameter, explicitly optimised for the current dataset. Secondly, we averaged the responses across trials; that is, we sampled the input from $\bar{x_d}$. The channels, randomly selected from $\bar{x_d}$, were processed by the network to generate a singular embedding. A multi-layer perceptron subsequently classified this embedding, as depicted in Figure \ref{fig:models}A.

\subsubsection[CNN (convergent)]{CNN\textsubscript{convergent}}
To facilitate a comparative analysis of both paradigms, we subsequently trained CNNs using a convergent paradigm. In this approach, we sample the model from $\bar{x_c}$, with the number of selected channels again set as a model hyperparameter. As discussed in section \ref{sec:intro}, this approach is potentially more suited for the task, given that stimulation from non-SOZ to SOZ sites results in larger responses than vice versa and that delayed responses occur at epileptogenic response sites. However, it was unclear if delayed responses would be detected, given that averaging across trials attenuates their prominence. Apart from these differences, the model architecture remained consistent with CNN\textsubscript{divergent}, as detailed in Figure \ref{fig:models}A.

\subsubsection[CNN Transformer]{CNN Transformer}
We next implemented encoder-only Transformer models with cross-channel attention, as introduced in Section \ref{sec:transformers}. We opt for a convergent paradigm, enabling a direct comparison with CNN\textsubscript{convergent}. Unlike the CNN models, the Transformer model does not require a fixed number of channels, thus allowing it to process responses from all channels. The network is depicted in Figure (Figure \ref{fig:models}B). For embeddings, a multi-scale 1D ResNet -- as with CNN\textsubscript{convergent} -- was employed; however, this process was conducted independently for each channel, leading to a distinct embedding per channel (Figure \ref{fig:models}C).

Next, a learnable [CLS] token is prepended to these
embeddings, and Transformer layers are applied with each channel treated as a
token. An MLP is attached to the final [CLS] representation to classify the input as SOZ or non-SOZ. This approach enables feature extraction across an arbitrary number of channels, ensuring its suitability for intracranial EEG.

\section{Experiments}
\label{sec:experiments}

\subsection{Dataset}
\label{sec:dataset}
For this study, we utilised an open-source dataset \citep{ds004080:1.2.4}, which comprises electrocorticography (ECoG) recordings from 74 patients aged between 4 and 51 years undergoing evaluation for epilepsy surgery\footnote{\url{https://openneuro.org/datasets/ds004080}}. We restricted our analyses to the 35 patients for whom each electrode was labelled as within or outside the SOZ. The ECoG data were recorded at a sampling rate of 2048 Hz. Stimulation was performed at 0.2 Hz with a current intensity of 4mA or 8mA. 

A total of 2066 ECoG electrodes were used, averaging 59 per patient. We restricted our analyses to stimulated electrodes to facilitate a fair comparison between both paradigms. Among these electrodes, 298 (14.4\%) were identified as within the SOZ. The proportion of electrodes in the SOZ was relatively consistent across lobes (Figure \ref{fig:lobe_distribution}). 

\begin{figure}[h]
\floatconts
  {fig:lobe_distribution}
  {\caption{\textbf{Electrode distribution across lobes.} This figure presents the percentage of all electrodes in each lobe versus those within the SOZ. It shows that the SOZ electrodes follow a similar distribution to all electrodes.}}
  {\includegraphics[width=0.75\linewidth]{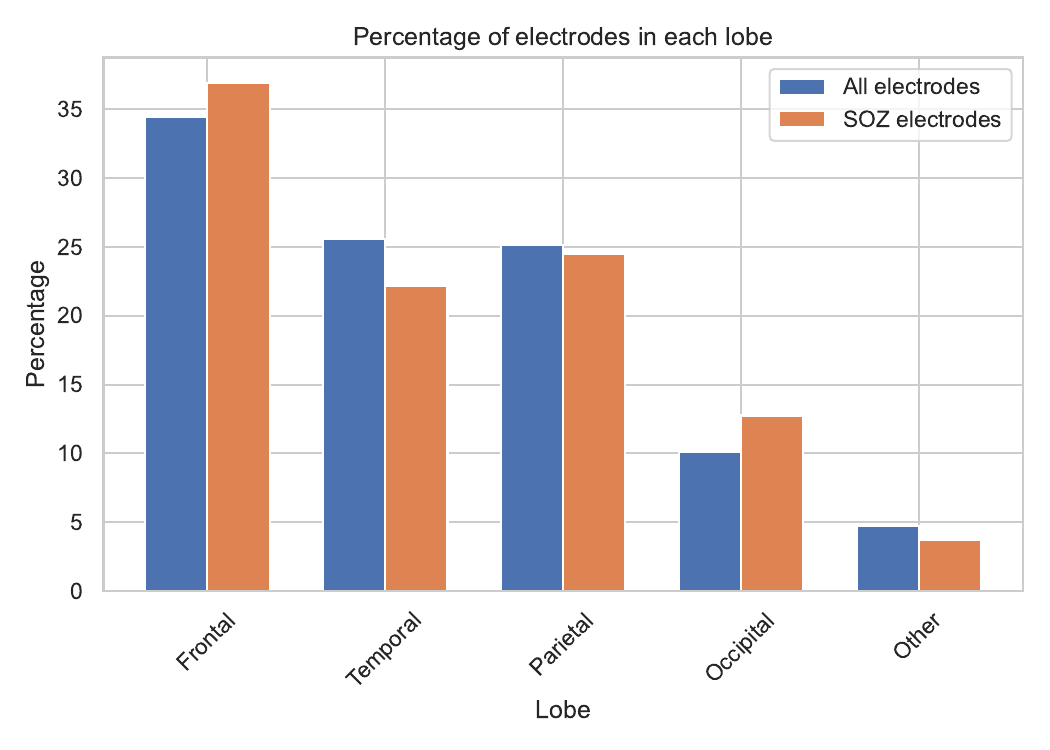}}
\end{figure}

\subsection{Dataset split}
We employed repeated k-fold cross-validation with $k=5$ and 5 repeats. To assess model generalisability to new patients and ECoG electrode locations, each patient was present in one fold, ensuring no patient data overlap between the training and testing sets. Further details are provided in Appendix \ref{ap:dataset_split}.

\subsection{Preprocessing}
Data were first bandpass filtered to 1-150 Hz in line with previous work in the same task \citep{johnson2022localizing}. Further preprocessing was applied in line with previous work using the same dataset \citep{van2023developmental}, as outlined in Appendix \ref{ap:preprocessing}. These preprocessed time series were used as input to each model, as outlined in section \ref{sec:methods}.

\subsection{Evaluation criteria}
To assess model performance across repeated k-fold cross-validation, we employed both threshold-independent and threshold-dependent metrics, each computed per patient and then averaged across patients, folds, and repeats. We used the Area Under the Receiver Operating Characteristic (AUROC) as the threshold-independent metric. We report the specificity, sensitivity, and Youden's index \citep{youden1950index} for threshold-dependent metrics, which consider binary rather than continuous predictions. We test whether the CNN Transformer demonstrates an improvement in performance compared to CNN\textsubscript{convergent}. The reported p-value assesses the statistical significance of the observed performance increase. Further details are included in Appendix \ref{ap:metrics}.

\subsection{Hyperparameter tuning}
We undertook a hyperparameter search for each model using Optuna \citep{akiba2019optuna} to ensure that observed differences in model performance were not due to inadequate hyperparameter configurations. This process led to the selection of 49 channels for CNN\textsubscript{divergent} and 37 channels for CNN\textsubscript{convergent}. Further details of the hyperparameter search spaces and the chosen hyperparameters for each model are in Appendix \ref{ap:model_training}.

\subsection{Channel sensitivity analysis}
Although the rationale for employing cross-channel attention is its potential to extract features from relevant channels more effectively, it is worth considering the extent to which the CNN Transformer's performance could be attributed to the availability of more channels during inference. If this were the case, it would indicate that the CNN Transformer benefits solely from handling arbitrary numbers of channels rather than the ability of cross-channel attention to extract relevant features. To investigate this aspect, we undertook a sensitivity analysis with the CNN Transformer, focusing on how its performance depends on the number of channels available during inference.

\subsection{Model performance by outcome}
An important consideration is whether the models offer complementary information to the SOZ. One possible mechanism for the models is that rather than approximating the SOZ -- a proxy for the epileptogenic zone (EZ) -- they may learn to approximate the EZ directly. Removal of the SOZ does not guarantee seizure freedom, so such a scenario could feasibly lead to improved outcomes. In general, we may expect that patients who became seizure-free had SOZs aligned closer with the EZ than those who continued to have seizures. Therefore, if the model approximates the EZ directly, we expect it to perform better in patients who became seizure-free. We tested this by performing an initial investigation to see the relationship between the CNN Transformer performance and surgical outcome. For this analysis, a modified version of the CNN Transformer that included the distance between stimulating and recording electrodes as a feature was used (mean AUROC: 0.745). We produced box plots of AUROC scores for seizure-free (Engel Class I) and non-seizure-free patients to investigate this. We then performed a one-tailed Mann-Whitney U test between these two groups to test for improved model performance in the seizure-free subcohort. 

\begin{table*}[!h]
\centering
\caption{\textbf{Model Performance Metrics.} Mean and standard deviation of the AUROC, AUPRC, specificity, sensitivity, and Youden's index for each model across all test folds and runs. \label{tab:results}}
\begin{tabular}{l|c|ccc}
\toprule
\bfseries Model & \bfseries AUROC & \bfseries Specificity & \bfseries Sensitivity & \bfseries Youden \\
\midrule
CNN\textsubscript{divergent} & $0.574 \pm 0.065$ & $0.632 \pm 0.118$ & $0.478 \pm 0.169$ & $0.109 \pm 0.106$ \\
CNN\textsubscript{convergent} & $0.666 \pm 0.067$ & $0.692 \pm 0.147$ & $0.531 \pm 0.199$ & $0.223 \pm 0.095$ \\
\midrule
CNN Transformer (ours) & $0.730 \pm 0.082$ & $0.731 \pm 0.130$ & $0.589 \pm 0.218$ & $0.319 \pm 0.134$ \\
\bottomrule
\end{tabular}
\end{table*}

\section{Results}
\label{sec:results}

\subsection{Performance metrics}


\begin{figure}[h]
\floatconts
  {fig:roc_curve_boxplot}
  {\caption{\textbf{A. ROC Curves.} Receiver Operating Characteristic (ROC) curves are shown for each model alongside a baseline random classifier. The AUROC scores are indicated for each model. \textbf{B. CNN Transformer performance by surgical outcome.} The per-patient AUROC scores were stratified by surgical outcome for the CNN Transformer. Scores were generally higher in the seizure-free cohort (median: 0.86) than in the non-seizure-free cohort (median: 0.77), suggesting that the models may predict outcomes. However, this difference was not statistically significant.}}
  {\includegraphics[width=\linewidth]{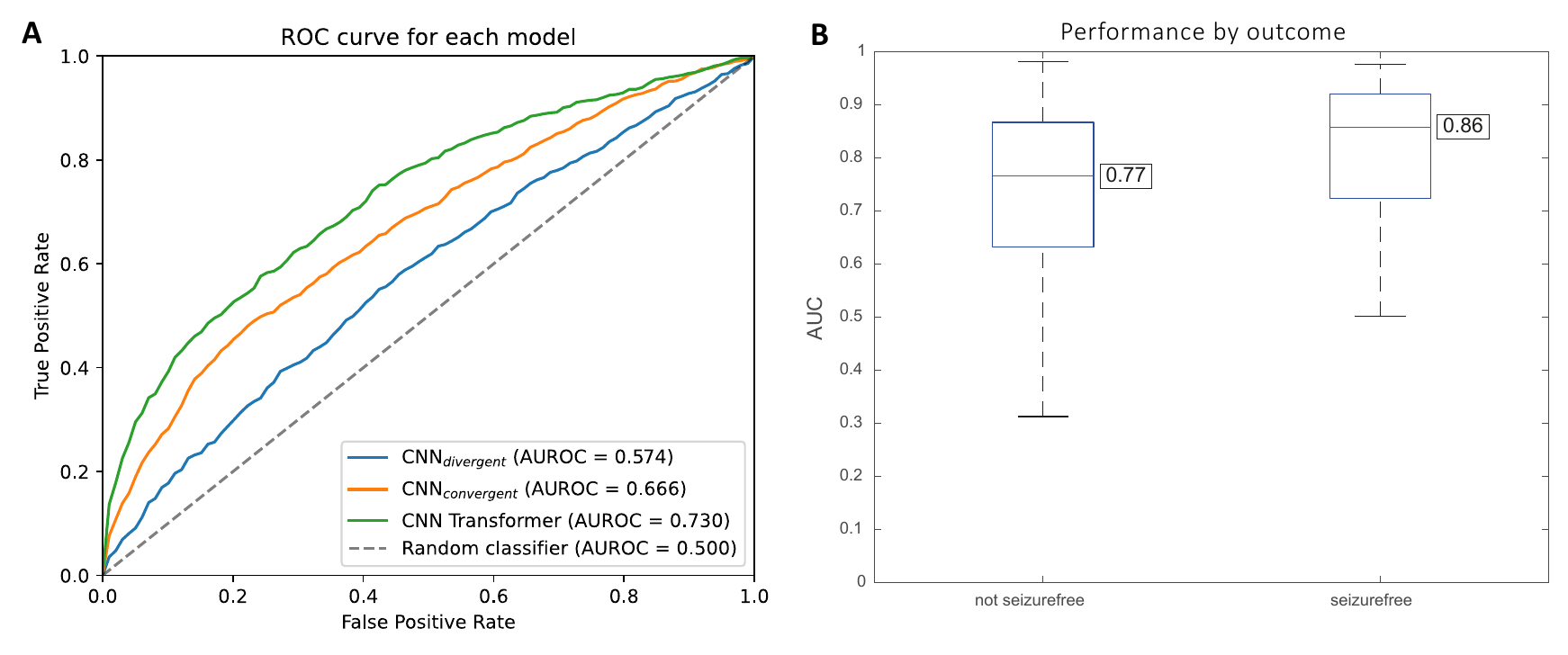}}
\end{figure}

Table \ref{tab:results} presents the performance metrics for each model. CNN\textsubscript{divergent}, which was previously shown to attain a Youden's index of 0.527 \citep{johnson2022localizing}, underperformed in this task, registering a Youden's index of only 0.109 and an AUROC marginally above chance at 0.574. In comparison, CNN\textsubscript{convergent} had an increased AUROC of 0.666, yet it still faced challenges in generalisation. When using cross-channel attention, the AUROC was 0.730. This improvement over CNN\textsubscript{convergent} was statistically significant at the 5\% level ($p=0.036$).

After selecting an optimal threshold, this hierarchy of model performance was maintained. The specificity and sensitivity values monotonically increased when switching to a convergent paradigm and then to a CNN Transformer. As the AUROC scores suggest, this hierarchy of model performance was consistent across thresholds (Figure \ref{fig:roc_curve_boxplot}A).

Figure \ref{fig:roc_curve_boxplot}B shows the distribution of AUROC scores for each outcome type. The general trend is that AUROC scores are higher in seizure-free patients following surgery ($n=11$), with a median AUROC of 0.86, compared to 0.77 in those not seizure-free ($n=23$). Furthermore, the lowest AUROC score was in a non-seizure-free patient. However, any improvement in AUROC was not statistically significant at the 5\% level ($p=0.115$).

\subsection{Channel sensitivity analysis}
\label{sec:channel_sensitivity}
The channel sensitivity analysis assessed the relationship between the performance of the CNN Transformer and the number of channels used during inference. A crucial finding was the model's robust performance across varying channel counts, including a monotonic increase in performance with an increasing number of channels, as illustrated in Figure \ref{fig:ablation}.

When the channel count was set to 37, in line with CNN\textsubscript{convergent}, our model achieved an AUROC of 0.724. This performance notably surpassed that of the CNN model, which had an AUROC of 0.666 with the same number of channels. Furthermore, the resilience of the CNN Transformer was underscored when the channel count was reduced to just 4, with an AUROC of 0.667, equalling that of CNN\textsubscript{convergent}.


\begin{figure}[h]
\floatconts
  {fig:ablation}
  {\caption{\textbf{Channel sensitivity analysis for CNN Transformer.} In this analysis, a random subset of $n$ channels served as input for each prediction, testing different values of $n$. This analysis used the previously trained CNN Transformer models. AUROC scores for CNN\textsubscript{convergent}, which used 37 channels, are presented for comparison. Bootstrapping was used to calculate 95\% confidence intervals.}}
  {\includegraphics[width=0.75\linewidth]{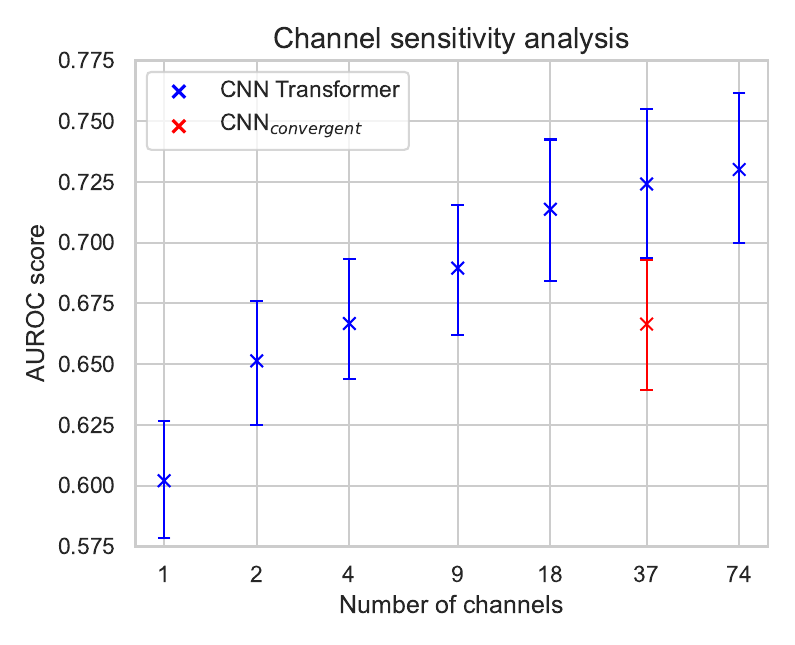}}
\end{figure}

\subsection{Predictions visualised}

Example predictions for a single patient are visualised in Figure \ref{fig:predictions}. Both CNN\textsubscript{divergent} and CNN\textsubscript{convergent} had a significant overlap with the clinician labelled SOZ, albeit with several false positives. The CNN Transformer exhibits a similar overlap but with fewer false positives.

\begin{figure}[h]
\floatconts
  {fig:predictions}
  {\caption{\textbf{Model outputs for a single patient.} (Left) Clinician-determined SOZ for an example patient. Nodes represent intracranial EEG electrodes. (Rest) Classification outputs for each model. Predictions from the CNN Transformer are most closely aligned with the clinician-determined SOZ. This patient was absent from the training set, demonstrating model generalisability to unseen patients and electrodes.}}
  {\includegraphics[width=\linewidth]{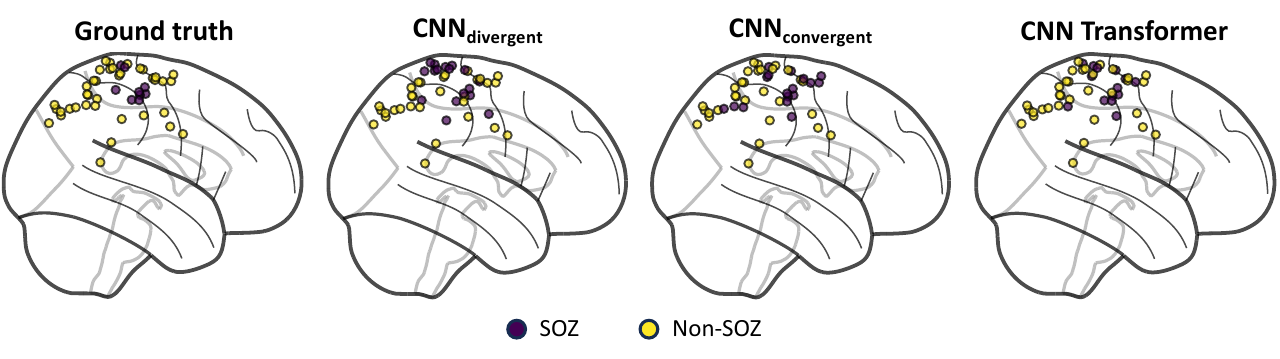}}
\end{figure}

\section{Discussion}
\subsection{Baseline performance}
\label{sec:baseline_discussion}

The change in dataset may have led to the suboptimal performance of CNN\textsubscript{divergent}. For instance, the stimulation parameters differed between the two datasets. \cite{johnson2022localizing} employed a stimulation frequency of 1 Hz, whereas \cite{ds004080:1.2.4} -- whose dataset we use -- applied a stimulation frequency of 0.2 Hz. It is plausible that the impaired performance can be attributed to the differences in responses elicited by stimulation at epileptogenic sites between these two frequencies.

The heterogeneity of seizure onsets in this dataset (see section \ref{sec:dataset}) compared with the previous focus on temporal lobe epilepsies may have been a contributing factor. Given the limited sample size, this may have led to underfitting. Another possible explanation is that our approach averaged responses across trials, attenuating delayed responses. However, \cite{johnson2022localizing} reported only a slight decrease in performance when limiting the post-stimulation window to 0-175ms -- i.e., removing most delayed responses, suggesting any performance reduction attributable to averaging across trials was likely negligible.

Finally, the dataset's composition, encompassing patients irrespective of treatment outcome, poses a significant challenge. Similar studies typically concentrate on patients with predominantly positive outcomes. Including patients who experienced unsuccessful treatment outcomes -- where the labelled SOZ may not accurately represent true epileptogenicity -- could systematically diminish the model's performance in attempting to infer actual epileptogenic regions.

Given the reduced performance of CNN\textsubscript{divergent} on the dataset employed in this study, it is desirable to compare our methods on the same dataset used by \cite{johnson2022localizing}. This approach would further validate the generalisability of our findings.

\subsection{Divergent vs Convergent paradigms}
In the introduction, we hypothesised that a \textit{convergent} paradigm might be more conducive to effective SOZ localisation. This hypothesis was premised on the literature suggesting that response sites are more indicative of epileptogenicity than stimulation sites. Our empirical investigation, which compared CNN\textsubscript{divergent} with CNN\textsubscript{convergent}, supports this hypothesis. The AUROC increased notably from 0.574 to 0.666, marking a substantial improvement, although this remains low for a task of critical importance.

\subsection{CNN vs CNN Transformer comparison}
In our comparative study of CNN\textsubscript{convergent} and the CNN Transformer, the latter demonstrated statistically significant superiority. An important question was whether this advantage was solely due to its ability to process all channels during inference rather than just a subset. Section \ref{sec:channel_sensitivity} shows evidence to the contrary, where the CNN Transformer outperformed CNN\textsubscript{convergent} even when restricted to the same number of inference channels.

The CNN Transformer's cross-channel attention mechanism enables improved feature extraction compared to CNN\textsubscript{convergent}, which applies a ResNet to compute a single set of features across channels. The superiority of the Transformer lies in its ability to extract channel-wise features and then apply cross-channel attention, explaining the performance disparity between the two models. Recent evidence has demonstrated the efficacy of cross-channel attention in intracranial EEG analysis \citep{zhang2024brant}. Our findings further extend its applicability, demonstrating its ability to model evoked responses effectively. Given the heterogeneity in patient-specific electrode configurations and its high spatial resolution, intracranial EEG is particularly well-suited to benefit from this mechanism.

\subsection{Model performance by outcome}
\label{sec:performance_by_outcome}
While the increase in AUROC in the seizure-free cohort may not have reached statistical significance, the emerging trend suggests that there is potential for a significant increase, given the right conditions:
\begin{enumerate}
    \item The small sample size ($n=35$) could be a limiting factor, as a larger cohort size is likely necessary to observe a significant increase in AUROC.
    \item The models were trained using data from patients with all outcomes. For patients who were not rendered seizure-free after surgery, some SOZ channels may not have been epileptogenic, and some non-SOZ channels may have been epileptogenic.
    \item The SOZ and resected channels typically diverge. A poor outcome does not imply that the SOZ did not indicate epileptogenicity, as removal of the SOZ may not have been possible.
\end{enumerate}
Whilst the overall Youden score of the CNN Transformer is only 0.319, which is likely some way off being clinically useful, the performance in the seizure-free cohort -- despite being trained on patients with all outcomes -- suggests the potential for improved clinical utility. A move towards predicting surgical outcomes is one such approach, in which the training data would be restricted to patients who went on to be seizure-free. Channels that were both SOZ and resected would be labelled as positive instances (epileptogenic), and those that were neither SOZ nor resected as negative instances (non-epileptogenic). We could then investigate whether resecting more of the highest scoring channels -- as determined by our models -- relates to surgical outcome.

\subsection{Clinical applications}
Our current models demonstrate significant potential to enhance clinical practice in epilepsy treatment by indicating the likelihood of an epileptogenic focus.

Firstly, we propose presenting the model’s outputs visually within a 3D spatial context, enabling clinicians to intuitively visualise abnormal brain regions. This visualisation can be integrated with multimodal pre-surgical evaluation data, such as radiological findings, seizure activity, interictal recordings, and PET imaging, to align with current clinical practice. In future, a multi-modal machine learning approach, utilising extensive, high-quality data, could significantly improve model performance, assuming complexity and scalability issues are overcome \citep{kline2022multimodal}.

Secondly, our model can flag areas where clinical labels and model predictions diverge, such as SOZ regions with low model scores or non-SOZ regions with high scores. Given the subjectivity in SOZ labelling, this feature can help clinicians identify additional regions of interest that may warrant further investigation, thus enhancing evaluation thoroughness beyond visual intracranial EEG analysis alone. Thirdly, providing continuous scores for each channel allows for a more nuanced assessment of epileptogenic activity likelihood. This detailed information supports clinicians in prioritising regions for surgical intervention and making more informed patient management decisions.

An alternative approach involves predicting surgical outcomes by restricting the training data to patients who became seizure-free post-surgery. Channels labelled as SOZ and resected would be positive instances (epileptogenic), while those neither SOZ nor resected would be negative instances (non-epileptogenic). This approach could help investigate whether resecting more of the highest-scoring channels, as determined by our models, correlates with better surgical outcomes. Further validation is crucial to confirm the effectiveness of these models in diverse patient populations and clinical settings, as discussed in subsequent sections.

\paragraph{Limitations}
A key challenge is the limited scope of our models' validation, which was limited to patients from the same dataset. Whilst we evaluated generalisability to unseen patients and electrode placements, our evaluation did not test for generalisability to other centres, which is particularly important given the variability in stimulation protocols and electrode modality (ECoG or SEEG). To validate their generalisability convincingly, testing on an external dataset is paramount. Ideally, such testing should be preceded by training models on a dataset derived from a multi-centre cohort, thus ensuring model robustness across clinical environments. Furthermore, we did not conduct a thorough analysis of the relationship between model performance and factors such as age, SOZ location, and electrode placement. Future validation studies on larger cohorts will be necessary to account for these potential effects.


A further divergence from clinical practice lies in the models' independently classifying each site. Intuitively, if one electrode site is deemed epileptogenic, we may also expect adjacent regions to be epileptogenic. Nonetheless, our current models do not account for this, as they make predictions for each electrode separately. Implementing post-processing techniques that consider the spatial relationship between electrodes could be beneficial in addressing this limitation.

Explainability is paramount in healthcare applications, where comprehending models' decision-making processes is critical. The reliance on complex `black box' models accentuates the necessity for explainable methodologies. Integrated gradients \citep{sundararajan2017axiomatic} represent one such approach. This method can verify that data artefacts do not improperly influence decisions by highlighting salient SPES responses. Such transparency is crucial for affirming model reliability and clinical applicability.

Finally, as discussed in section \ref{sec:intro}, various forms of inter-trial variability can indicate epileptogenicity. Averaging across trials increased the signal-to-noise ratio of SPES responses at the expense of these markers. Several approaches can be used to factor in this variability, such as capturing the standard deviation across trials. Alternatively, cross-trial attention can be used to directly extract features from relevant trials. This approach may be particularly suited to detecting delayed responses, which predict surgical outcome \citep{valentin2005single}.

\section{Conclusion}
\label{sec:conclusion}

This paper presents several contributions for modelling intracranial EEG data using deep learning, particularly in localising the SOZ from SPES data. Our research first compares \textit{divergent} and \textit{convergent} paradigms in classifying multi-channel SPES responses. This comparison revealed a substantial improvement in moving from the conventional divergent approach (AUROC: 0.574) to a convergent one (AUROC: 0.666). Our approach of applying a Transformer with cross-channel attention to invasive EEG data has shown promising results, elevating the AUROC to 0.730.

We aim to extend our work in several directions. One key area is investigating whether our approach can predict epileptogenicity more accurately than traditional SOZ localisation. There is currently only limited evidence for this, as any improvement in performance in the seizure-free cohort did not reach statistical significance. Another avenue is to explore the potential of the CNN Transformer to process SPES responses across trials rather than taking the mean. Additionally, evaluating our model on an external dataset will be crucial, given the heterogeneity in SPES protocols.

Our study significantly advances the modelling of heterogeneous intracranial EEG placements in SPES. We hope that further developments can improve surgical outcomes for patients with drug-resistant epilepsy. 
    
\acks{JN is supported by a UCL UKRI Centre for Doctoral Training in AI-enabled Healthcare studentship (EP/S021612/1).}

\bibliography{sample}

\begin{thebibliography}{32}
\providecommand{\natexlab}[1]{#1}
\providecommand{\url}[1]{\texttt{#1}}
\expandafter\ifx\csname urlstyle\endcsname\relax
  \providecommand{\doi}[1]{doi: #1}\else
  \providecommand{\doi}{doi: \begingroup \urlstyle{rm}\Url}\fi

\bibitem[Akiba et~al.(2019)Akiba, Sano, Yanase, Ohta, and Koyama]{akiba2019optuna}
Takuya Akiba, Shotaro Sano, Toshihiko Yanase, Takeru Ohta, and Masanori Koyama.
\newblock Optuna: A next-generation hyperparameter optimization framework.
\newblock In \emph{Proceedings of the 25th ACM SIGKDD international conference on knowledge discovery \& data mining}, pages 2623--2631, 2019.

\bibitem[Bahdanau et~al.(2014)Bahdanau, Cho, and Bengio]{bahdanau2014neural}
Dzmitry Bahdanau, Kyunghyun Cho, and Yoshua Bengio.
\newblock Neural machine translation by jointly learning to align and translate.
\newblock \emph{arXiv preprint arXiv:1409.0473}, 2014.

\bibitem[Chen et~al.(2021)Chen, Lu, Rajeswaran, Lee, Grover, Laskin, Abbeel, Srinivas, and Mordatch]{chen2021decision}
Lili Chen, Kevin Lu, Aravind Rajeswaran, Kimin Lee, Aditya Grover, Misha Laskin, Pieter Abbeel, Aravind Srinivas, and Igor Mordatch.
\newblock Decision transformer: Reinforcement learning via sequence modeling.
\newblock \emph{Advances in neural information processing systems}, 34:\penalty0 15084--15097, 2021.

\bibitem[Cornblath et~al.(2023)Cornblath, Lucas, Armstrong, Greenblatt, Stein, Hadar, Raghupathi, Marsh, Litt, Davis, et~al.]{cornblath2023quantifying}
Eli~J Cornblath, Alfredo Lucas, Caren Armstrong, Adam~S Greenblatt, Joel~M Stein, Peter~N Hadar, Ramya Raghupathi, Eric Marsh, Brian Litt, Kathryn~A Davis, et~al.
\newblock Quantifying trial-by-trial variability during cortico-cortical evoked potential mapping of epileptogenic tissue.
\newblock \emph{Epilepsia}, 2023.

\bibitem[Devlin et~al.(2018)Devlin, Chang, Lee, and Toutanova]{devlin2018bert}
Jacob Devlin, Ming-Wei Chang, Kenton Lee, and Kristina Toutanova.
\newblock Bert: Pre-training of deep bidirectional transformers for language understanding.
\newblock \emph{arXiv preprint arXiv:1810.04805}, 2018.

\bibitem[Dosovitskiy et~al.(2020)Dosovitskiy, Beyer, Kolesnikov, Weissenborn, Zhai, Unterthiner, Dehghani, Minderer, Heigold, Gelly, et~al.]{dosovitskiy2020image}
Alexey Dosovitskiy, Lucas Beyer, Alexander Kolesnikov, Dirk Weissenborn, Xiaohua Zhai, Thomas Unterthiner, Mostafa Dehghani, Matthias Minderer, Georg Heigold, Sylvain Gelly, et~al.
\newblock An image is worth 16x16 words: Transformers for image recognition at scale.
\newblock \emph{arXiv preprint arXiv:2010.11929}, 2020.

\bibitem[Hays et~al.(2023)Hays, Smith, Wang, Coogan, Sarma, Crone, and Kang]{hays2023cortico}
Mark~A Hays, Rachel~J Smith, Yujing Wang, Christopher Coogan, Sridevi~V Sarma, Nathan~E Crone, and Joon~Y Kang.
\newblock Cortico-cortical evoked potentials in response to varying stimulation intensity improves seizure localization.
\newblock \emph{Clinical Neurophysiology}, 145:\penalty0 119--128, 2023.

\bibitem[He et~al.(2016)He, Zhang, Ren, and Sun]{he2016deep}
Kaiming He, Xiangyu Zhang, Shaoqing Ren, and Jian Sun.
\newblock Deep residual learning for image recognition.
\newblock In \emph{Proceedings of the IEEE conference on computer vision and pattern recognition}, pages 770--778, 2016.

\bibitem[Hussein et~al.(2022)Hussein, Lee, and Ward]{hussein2022multi}
Ramy Hussein, Soojin Lee, and Rabab Ward.
\newblock Multi-channel vision transformer for epileptic seizure prediction.
\newblock \emph{Biomedicines}, 10\penalty0 (7):\penalty0 1551, 2022.

\bibitem[Jehi(2018)]{jehi2018epileptogenic}
Lara Jehi.
\newblock The epileptogenic zone: concept and definition.
\newblock \emph{Epilepsy currents}, 18\penalty0 (1):\penalty0 12--16, 2018.

\bibitem[Johnson et~al.(2022)Johnson, Cai, Doss, Jiang, Negi, Narasimhan, Paulo, Gonz{\'a}lez, Roberson, Bick, et~al.]{johnson2022localizing}
Graham~W Johnson, Leon~Y Cai, Derek~J Doss, Jasmine~W Jiang, Aarushi~S Negi, Saramati Narasimhan, Danika~L Paulo, Hern{\'a}n~FJ Gonz{\'a}lez, Shawniqua~Williams Roberson, Sarah~K Bick, et~al.
\newblock Localizing seizure onset zones in surgical epilepsy with neurostimulation deep learning.
\newblock \emph{Journal of Neurosurgery}, 138\penalty0 (4):\penalty0 1002--1007, 2022.

\bibitem[Kalilani et~al.(2018)Kalilani, Sun, Pelgrims, Noack-Rink, and Villanueva]{kalilani2018epidemiology}
Linda Kalilani, Xuezheng Sun, Barbara Pelgrims, Matthias Noack-Rink, and Vicente Villanueva.
\newblock The epidemiology of drug-resistant epilepsy: a systematic review and meta-analysis.
\newblock \emph{Epilepsia}, 59\penalty0 (12):\penalty0 2179--2193, 2018.

\bibitem[Kline et~al.(2022)Kline, Wang, Li, Dennis, Hutch, Xu, Wang, Cheng, and Luo]{kline2022multimodal}
Adrienne Kline, Hanyin Wang, Yikuan Li, Saya Dennis, Meghan Hutch, Zhenxing Xu, Fei Wang, Feixiong Cheng, and Yuan Luo.
\newblock Multimodal machine learning in precision health.
\newblock \emph{arXiv preprint arXiv:2204.04777}, 2022.

\bibitem[Lim et~al.(2021)Lim, Ar{\i}k, Loeff, and Pfister]{lim2021temporal}
Bryan Lim, Sercan~{\"O} Ar{\i}k, Nicolas Loeff, and Tomas Pfister.
\newblock Temporal fusion transformers for interpretable multi-horizon time series forecasting.
\newblock \emph{International Journal of Forecasting}, 37\penalty0 (4):\penalty0 1748--1764, 2021.

\bibitem[Loshchilov and Hutter(2017)]{loshchilov2017decoupled}
Ilya Loshchilov and Frank Hutter.
\newblock Decoupled weight decay regularization.
\newblock \emph{arXiv preprint arXiv:1711.05101}, 2017.

\bibitem[Malone et~al.(2022)Malone, Smith, Urdaneta, Davis, Anderson, Phillip, Rolston, and Butson]{malone2022machine}
Ian~G Malone, Kaleb~E Smith, Morgan~E Urdaneta, Tyler~S Davis, Daria~Nesterovich Anderson, Brian~J Phillip, John~D Rolston, and Christopher~R Butson.
\newblock Machine learning methods applied to cortico-cortical evoked potentials aid in localizing seizure onset zones.
\newblock \emph{arXiv preprint arXiv:2211.07867}, 2022.

\bibitem[Matsumoto et~al.(2017)Matsumoto, Kunieda, and Nair]{matsumoto2017single}
Riki Matsumoto, Takeharu Kunieda, and Dileep Nair.
\newblock Single pulse electrical stimulation to probe functional and pathological connectivity in epilepsy.
\newblock \emph{Seizure}, 44:\penalty0 27--36, 2017.

\bibitem[Miller et~al.(2021)Miller, M{\"u}ller, and Hermes]{miller2021basis}
Kai~J Miller, Klaus-Robert M{\"u}ller, and Dora Hermes.
\newblock Basis profile curve identification to understand electrical stimulation effects in human brain networks.
\newblock \emph{PLoS computational biology}, 17\penalty0 (9):\penalty0 e1008710, 2021.

\bibitem[Nadeau and Bengio(1999)]{nadeau1999inference}
Claude Nadeau and Yoshua Bengio.
\newblock Inference for the generalization error.
\newblock \emph{Advances in neural information processing systems}, 12, 1999.

\bibitem[Paszke et~al.(2019)Paszke, Gross, Massa, Lerer, Bradbury, Chanan, Killeen, Lin, Gimelshein, Antiga, et~al.]{paszke2019pytorch}
Adam Paszke, Sam Gross, Francisco Massa, Adam Lerer, James Bradbury, Gregory Chanan, Trevor Killeen, Zeming Lin, Natalia Gimelshein, Luca Antiga, et~al.
\newblock Pytorch: An imperative style, high-performance deep learning library.
\newblock \emph{Advances in neural information processing systems}, 32, 2019.

\bibitem[Sundararajan et~al.(2017)Sundararajan, Taly, and Yan]{sundararajan2017axiomatic}
Mukund Sundararajan, Ankur Taly, and Qiqi Yan.
\newblock Axiomatic attribution for deep networks.
\newblock In \emph{International conference on machine learning}, pages 3319--3328. PMLR, 2017.

\bibitem[{UK Children’s Epilepsy Surgery Collaboration} et~al.(2021){UK Children’s Epilepsy Surgery Collaboration}, Chari, Moeller, Boyd, Tahir, Cross, Eltze, Das, van Dalen, Scott, et~al.]{uk2021uk}
{UK Children’s Epilepsy Surgery Collaboration}, Aswin Chari, Friederike Moeller, Stewart Boyd, M~Zubair Tahir, J~Helen Cross, Christin Eltze, Krishna Das, Thijs van Dalen, Rod~C Scott, et~al.
\newblock The uk experience of stereoelectroencephalography in children: An analysis of factors predicting the identification of a seizure-onset zone and subsequent seizure freedom.
\newblock \emph{Epilepsia}, 62\penalty0 (8):\penalty0 1883--1896, 2021.

\bibitem[Valent{\'\i}n et~al.(2005)Valent{\'\i}n, Alarc{\'o}n, Honavar, Seoane, Selway, Polkey, and Binnie]{valentin2005single}
Antonio Valent{\'\i}n, Gonzalo Alarc{\'o}n, Mrinalini Honavar, Jorge J~Garc{\'\i}a Seoane, Richard~P Selway, Charles~E Polkey, and Colin~D Binnie.
\newblock Single pulse electrical stimulation for identification of structural abnormalities and prediction of seizure outcome after epilepsy surgery: a prospective study.
\newblock \emph{The Lancet Neurology}, 4\penalty0 (11):\penalty0 718--726, 2005.

\bibitem[van Blooijs et~al.(2023{\natexlab{a}})van Blooijs, van~den Boom, van~der Aar, Huiskamp, Castegnaro, Demuru, Zweiphenning, van Eijsden, Miller, Leijten, and Hermes]{ds004080:1.2.4}
D.~van Blooijs, M.A. van~den Boom, J.F. van~der Aar, G.J.M. Huiskamp, G.~Castegnaro, M.~Demuru, W.J.E.M. Zweiphenning, P.~van Eijsden, K.~J. Miller, F.S.S. Leijten, and D.~Hermes.
\newblock "ccep ecog dataset across age 4-51", 2023{\natexlab{a}}.

\bibitem[van Blooijs et~al.(2023{\natexlab{b}})van Blooijs, van~den Boom, van~der Aar, Huiskamp, Castegnaro, Demuru, Zweiphenning, van Eijsden, Miller, Leijten, et~al.]{van2023developmental}
Dorien van Blooijs, Max~A van~den Boom, Jaap~F van~der Aar, Geertjan~M Huiskamp, Giulio Castegnaro, Matteo Demuru, Willemiek~JEM Zweiphenning, Pieter van Eijsden, Kai~J Miller, Frans~SS Leijten, et~al.
\newblock Developmental trajectory of transmission speed in the human brain.
\newblock \emph{Nature Neuroscience}, 26\penalty0 (4):\penalty0 537--541, 2023{\natexlab{b}}.

\bibitem[Vaswani et~al.(2017)Vaswani, Shazeer, Parmar, Uszkoreit, Jones, Gomez, Kaiser, and Polosukhin]{vaswani2017attention}
Ashish Vaswani, Noam Shazeer, Niki Parmar, Jakob Uszkoreit, Llion Jones, Aidan~N Gomez, {\L}ukasz Kaiser, and Illia Polosukhin.
\newblock Attention is all you need.
\newblock \emph{Advances in neural information processing systems}, 30, 2017.

\bibitem[Vaughan et~al.(2018)Vaughan, Ramos, Buch, Mekary, Amundson, Shah, Rattani, Dewan, and Park]{vaughan2018estimation}
Kerry~A Vaughan, Christian~Lopez Ramos, Vivek~P Buch, Rania~A Mekary, Julia~R Amundson, Meghal Shah, Abbas Rattani, Michael~C Dewan, and Kee~B Park.
\newblock An estimation of global volume of surgically treatable epilepsy based on a systematic review and meta-analysis of epilepsy.
\newblock \emph{Journal of neurosurgery}, 130\penalty0 (4):\penalty0 1127--1141, 2018.

\bibitem[Wang et~al.(2018)Wang, Han, Zhang, He, and Huang]{wang2018csi}
Fei Wang, Jinsong Han, Shiyuan Zhang, Xu~He, and Dong Huang.
\newblock Csi-net: Unified body characterization and action recognition.
\newblock \emph{arXiv preprint arXiv:1810.03064}, 2018.

\bibitem[Wiebe et~al.(2001)Wiebe, Blume, Girvin, and Eliasziw]{wiebe2001randomized}
Samuel Wiebe, Warren~T Blume, John~P Girvin, and Michael Eliasziw.
\newblock A randomized, controlled trial of surgery for temporal-lobe epilepsy.
\newblock \emph{New England Journal of Medicine}, 345\penalty0 (5):\penalty0 311--318, 2001.

\bibitem[Yang et~al.(2024)Yang, Zhao, Li, Mo, Guo, Li, Yao, Fan, Cai, Sang, et~al.]{yang2024localizing}
Bowen Yang, Baotian Zhao, Chao Li, Jiajie Mo, Zhihao Guo, Zilin Li, Yuan Yao, Xiuliang Fan, Du~Cai, Lin Sang, et~al.
\newblock Localizing seizure onset zone by a cortico-cortical evoked potentials-based machine learning approach in focal epilepsy.
\newblock \emph{Clinical Neurophysiology}, 2024.

\bibitem[Youden(1950)]{youden1950index}
William~J Youden.
\newblock Index for rating diagnostic tests.
\newblock \emph{Cancer}, 3\penalty0 (1):\penalty0 32--35, 1950.

\bibitem[Zhang et~al.(2024)Zhang, Yuan, Yang, Chen, Wang, and Li]{zhang2024brant}
Daoze Zhang, Zhizhang Yuan, Yang Yang, Junru Chen, Jingjing Wang, and Yafeng Li.
\newblock Brant: Foundation model for intracranial neural signal.
\newblock \emph{Advances in Neural Information Processing Systems}, 36, 2024.

\end{thebibliography}

\newpage
\appendix

\section{Dataset split}
\label{ap:dataset_split}
We repeated the following five times, each with a different random seed. We divided the dataset at the patient level into $k=5$ folds, such that all data for a given patient was in one fold only. In every run, we used three of these folds for training, one for validation, and one for testing. This approach was chosen to balance computational resource demands and the need for reliable model evaluation.

\section{Preprocessing}
\label{ap:preprocessing}
Firstly, we corrected each stimulation epoch for baseline by removing the mean from a preceding 900 ms window (-1 s to -0.1 s). As stimulation artefact can spread through volume conductance effects, we removed the first 9 ms post-stimulation and any electrodes within 13 mm of the stimulated electrode. As known responses to SPES mainly occur within 1 s, this window was maintained such that each epoch was from 9 ms to 1 s post-stimulation.

After dividing the dataset into training, validation, and test sets for each run, we carried out standardisation to ensure zero mean and unit standard deviation. This process involved computing the mean and standard deviation across the training set. We used these calculated parameters from the training set to independently standardise the training, validation, and test sets to prevent data leakage.

During training, we added noise with zero mean and a standard deviation of 0.1 for data augmentation. We removed a subset of channels during each training pass for the Transformer models, with a proportion uniformly distributed from [0, 0.5].

\section{Metrics}
\label{ap:metrics}
The AUROC score considers class imbalance, with a score of 0.5 corresponding to a random classifier and 1.0 to a perfect classifier. This metric indicates, at the patient level, whether SOZ electrodes are generally scored higher than non-SOZ electrodes.

A threshold was required to compute the specificity, sensitivity, and Youden's index. We determined patient-specific thresholds in the following manner:
\begin{enumerate}
    \item For each patient in the validation set, we collated model predictions (SOZ likelihood) for each electrode ($x$).
    \item We determined the patient-specific threshold that maximised the Youden score ($thresh$).
    \item We calculated $n$ which satisfied the following: $thresh = median(x) + n\times MAD(x) $, where $MAD$ is the median absolute deviation.
    \item We calculated the median $n$ across patients in the validation set.
\end{enumerate}
This allowed us to calculate a patient-specific threshold for each patient in the test set that required no knowledge of which channels were SOZ.

The statistical test between CNN\textsubscript{convergent} and the CNN Transformer was a one-sided t-test using \cite{nadeau1999inference}'s corrected method for repeated k-fold cross-validation. We calculated the p-value based on the mean AUROC scores from each fold and run, with a significance threshold set at 0.05.

\section{Model training procedures}
\label{ap:model_training}
Models were trained in PyTorch \citep{paszke2019pytorch}. An AdamW optimiser was used for training \citep{loshchilov2017decoupled} with a binary cross-entropy loss function. The \textit{pos\_weight} argument was used to account for class imbalance. The default parameters were used, except for the learning rate, which was determined through the hyperparameter search. For the CNN models, the search space was as follows: (1) learning rate within [$1 \times 10^{-4}$, $1 \times 10^{-2}$], (2) dropout rate at the fully connected layer within [0, 0.5], (3) number of input trials, an integer from [20, 80]. For the CNN Transformer, this search space was: (1) learning rate within [$1 \times 10^{-4}$, $1 \times 10^{-2}$], (2) dropout rate after the embedding layer, in the Transformer layers, and at the fully connected layer within [0, 0.5], (3) embedding dimension from \{16, 32, 64\}, (4) Transformer layers from \{1, 2\}. The number of Transformer heads was set to 1/8 of the embedding dimension. These search spaces were determined based on initial experiments on the validation sets. 

Ten trials were conducted for each model to balance model performance against computational resource constraints. Training was halted after ten epochs, and the set of hyperparameters that maximised the validation AUROC were selected. This process was only done once, with a separate random seed to the 5 runs, as this process took several times longer than the final model training. The best hyperparameters for the CNN and CNN Transformer models are shown in Tables \ref{tab:cnn_hyperparams} and \ref{tab:transformer_hyperparams}, respectively. 

\begin{table*}[h]
\centering
\caption{\textbf{Hyperparameters for CNN Models.} The table displays the optimal dropout rate and input channels for both CNN models as determined by the hyperparameter search. \label{tab:cnn_hyperparams}}
\begin{tabular}{l|ccc}
\toprule
\bfseries Model & \bfseries Dropout Rate & \bfseries Input Channels & \bfseries Learning rate\\
\midrule
CNN\textsubscript{divergent} & 0.22 & 49 & 4.0 $\times 10^{-3}$ \\
CNN\textsubscript{convergent} & 0.44 & 37 & 1.3 $\times 10^{-3}$ \\
\bottomrule
\end{tabular}
\end{table*}

\begin{table*}[h]
\centering
\caption{\textbf{Hyperparameters for the CNN Transformer.} The table presents the optimal dropout rate, embedding dimension, and number of layers for the CNN Transformer. \label{tab:transformer_hyperparams}}
\begin{tabular}{l|cccc}
\toprule
\bfseries Model & \bfseries Dropout Rate & \bfseries Embedding Dim & \bfseries Num Layers & \bfseries Learning rate \\
\midrule
CNN Transformer & 0.46 & 16 & 2 & 1.5 $\times 10^{-4}$ \\
\bottomrule
\end{tabular}
\end{table*}

\end{document}